\documentclass[journal]{IEEEtran}

\usepackage{amsmath,amssymb,amsfonts}
\usepackage{algorithmic,microtype}
\usepackage{dsfont}
\usepackage{textcomp}
\usepackage{xcolor}

\ifCLASSINFOpdf
\else
  \usepackage[dvips]{graphicx}
\fi
\usepackage{url}

\usepackage{graphicx}

\ifCLASSOPTIONcompsoc
    \usepackage[caption=false, font=normalsize, labelfont=sf, textfont=sf]{subfig}
\else
\usepackage[caption=false, font=footnotesize]{subfig}
\fi

\usepackage[2]{editing} 
\DeclareMathAlphabet\calbf{OMS}{cmsy}{b}{n}
\begin{document}

\title{Energy-Efficient Distributed Learning Algorithms for Coarsely Quantized Signals \vspace{-0.35em}}

\author{Alireza Danaee, Rodrigo C. de Lamare and Vítor H. Nascimento \vspace{-2.0em}
\thanks{Alireza Danaee and Rodrigo C. de Lamare are with CETUC, Pontifical Catholic University of Rio de Janeiro, Rio de Janeiro 22451-900, Brazil (\{alireza, delamare\}@cetuc.puc-rio.br). Vítor H. Nascimento is with the Department of Electronic Systems Engineering, University of São Paulo, Brazil (vitor@lps.usp.br). This work is funded in part by FAPESP Project 2018/12579-7 (ELIOT).}}

\markboth{IEEE SIGNAL PROCESSING LETTERS, Vol. X, No. X, MM YYYY}
{Shell \MakeLowercase{\textit{et al.}}: Bare Demo of IEEEtran.cls for IEEE Journals}
\maketitle

\begin{abstract}
In this work, we present an energy-efficient distributed learning framework using low-resolution ADCs and coarsely quantized signals for Internet of Things (IoT) networks. In particular, we develop a distributed quantization-aware least-mean square (DQA-LMS) algorithm that can learn parameters in an energy-efficient fashion using signals quantized with few bits while requiring a low computational cost. We also carry out a statistical analysis of the proposed DQA-LMS algorithm that includes a stability condition. Simulations assess the DQA-LMS algorithm against existing techniques for a distributed parameter estimation task where IoT devices operate in a peer-to-peer mode and demonstrate the effectiveness of the DQA-LMS algorithm.
\end{abstract}
\vspace{-0.25em}


\IEEEpeerreviewmaketitle
\vspace{-0.25em}
\section{Introduction}

\IEEEPARstart{D}{istributed} signal processing algorithms are of great relevance for statistical inference in wireless networks and applications such as wireless sensor networks (WSNs) \cite{predd2006distributed} and the Internet of Things (IoT) \cite{rana2018iot}. These techniques deal with the extraction of information from data collected at nodes that are distributed over a geographic area.
Prior work on distributed approaches has studied protocols for exchanging information \cite{olfati2007,lopes2008diffusion,sayed2013diffusion}, adaptive learning algorithms \cite{jio,jidf,xu2016distributed,miller2016distributed}, the exploitation of sparse and low-rank measurements \cite{dce,miller2015sparsity,djio}, topology adaptation \cite{xu2015adaptive}, compensation methods for highly correlated input signals \cite{zhang2020distributed}, and robust techniques against interference and noise \cite{yu2019robust}. Although there are many studies on the need for data exchange and signaling among nodes as well as their complexity, prior work on energy-efficient techniques is rather limited.

In this context, energy-efficient signal processing techniques have gained a great deal of interest in the last decade or so due to their ability to save energy and promote sustainable development of electronic systems and devices. Electronic devices often exhibit a power consumption that is dependent on the communication module \cite{han2013cross,utlu2017resource} and from a circuit perspective on analog-to-digital converters (ADCs) and  decoders \cite{mezghani2011power}. Reducing the number of bits used to represent digital samples can greatly decrease the energy consumption by ADCs \cite{walden1999analog}. This is key to devices that are battery operated and wireless networks that must keep the power consumption to a low level for sustainability reasons. In particular, prior work on energy efficiency has reported many contributions in signal processing for communications and electronic systems that operate with coarsely quantized signals \cite{bbprec2017,jacobsson2017throughput,mezghani2007modified,1bitcpm,1bitidd,shao2018adaptive,1bitce,dynover}.

In this work, we propose an energy-efficient distributed learning framework using low-resolution ADCs and coarsely quantized signals for IoT networks \cite{dqaspl}. In particular, we devise a distributed quantization-aware least-mean square (DQA-LMS) algorithm that can learn parameters in an energy-efficient way using signals quantized using few bits with a low computational cost. We also develop a statistical analysis of the DQA-LMS algorithm that includes a stability condition. Simulations assess the DQA-LMS algorithm against existing techniques for a distributed parameter estimation task with IoT devices. % where IoT devices operate in a peer-to-peer mode.

This paper is structured as follows: Section II introduces the signal model and states the problem. Section III details the proposed DQA-LMS algorithm, whereas Section IV analyzes DQA-LMS. Section V shows and discusses the simulation results and Section VI draws the conclusions of this work.
\vspace{-0.25em}
\section{Signal Model and Problem Statement}
\label{sec:syst}
\vspace{-1.25em}
\begin{figure}[htbp]
    \centering
    \includegraphics[width=6.9cm]{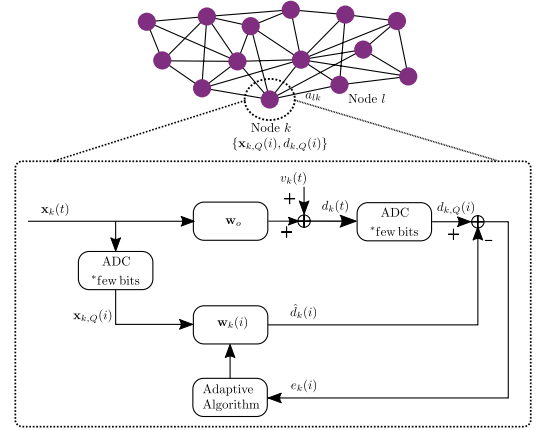}
    \vspace{-0.75em}
    \caption{A distributed adaptive IoT network}
    \label{dnet}
\end{figure}
\vspace{-0.5em}

We consider an IoT network consisting of $N$ nodes or agents, which run distributed signal processing techniques to perform the desired tasks, as depicted in Fig.~\ref{dnet}. The model adopted considers a desired signal $d_{k}(i)$, at each time $i$, described by 
\begin{equation}
        d_k(i) = {\mathbf w}_o^H {\mathbf x}_{k}(i)+v_{k}(i), \quad k=1,2,\dots,N  \label{regres},
\end{equation}
where ${\mathbf w}_o \in \mathbb{C}^{M \times 1}$ is the parameter vector that the agents must estimate, ${\mathbf x}_k(i) \in \mathbb{C}^{M \times 1}$ is the regressor and $v_{k}(i)$ represents Gaussian noise with zero mean and variance $\sigma_{v,k}^2$ at node $k$. We adopt the Adapt-then-Combine (ATC) diffusion rule as it outperforms the incremental and consensus protocols \cite{olfati2007,lopes2008diffusion}. At each node $k$ and time $i$, based on the local data \{$d_k(i)$, ${\mathbf x}_k(i)$\} and the estimated parameter vectors ${\mathbf h}_{l}(i)$ from its neighborhood, the parameter vector with local estimates ${\mathbf w}_k(i)$ is updated. The ATC distributed LMS (DLMS) algorithm consists of the recursions:
\begin{align*}
    %\begin{split}
        {\mathbf h}_{k}(i)=&{\mathbf w}_{k}(i-1)+\mu _{k}{\mathbf x}_{k}(i)e_k^*(i),  &
        {\mathbf w}_{k}(i)=&\sum _{l\in \mathcal {N}_{k}}a_{lk}{\mathbf h}_{l}(i), \label{atcw}
    %\end{split}
\end{align*}
where ${\mathbf h}_{k}(i)$ and ${\mathbf w}_{k}(i)$ contain the intermediate and the local estimates of ${\mathbf w}_o$ at node $k$ and time $i$, respectively, $e_k(i)=d_{k}(i)-\hat{d}_k(i)=d_{k}(i)-{\mathbf w}_k^H(i-1){\mathbf x}_k(i)$ is the error between the output of the adaptive filter, $\hat{d}_k(i)$, and the desired signal, $d_k(i)$, at time $i$, $\mu_{k}$ is the step-size for node $k$, $\mathcal {N}_{k}$ is the set of neighbor nodes connected to node $k$, and $a_{lk}$ are the combination coefficients of neighbor nodes at node $k$ such that
\begin{equation}
    a_{lk}=0 \, ~{\rm if} \, l \notin \mathcal {N}_k,~ a_{lk}>0 \,~ {\rm if} \, l \in \mathcal {N}_k, ~{\rm and} \sum_{l\in{\mathcal{N}_k}} {a_{lk}} = 1 \label{alk}.
    \raisetag{10pt}
\end{equation}
As shown in Fig.~\ref{dnet}, as the measurement data at each node and the unknown system are analog and each agent processes local data \{$d_k(i)$, ${\mathbf x}_k(i)$\} digitally, we need two ADCs in each agent. One concern is that as the number of agents increases, the power consumption will grow considerably when using high-resolution ADCs for each agent. This motivates us to quantize signals using few bits. Therefore, the problem we are interested in solving is how to design energy-efficient distributed learning algorithms that can cost-effectively operate with coarsely quantized signals. \vspace{-0.5em}

\section{Proposed DQA-LMS Algorithm}
\label{sec:proposed}
Let ${\mathbf x}_{k,Q}=Q_b({\mathbf x}_k)$ denote the $b$-bit quantized output of an ADC at node $k$, described by a set of $2^b+1$ thresholds ${\cal T}_b=\{\tau_0,\tau_1,...,\tau_{2^b}\}$, such that $-\infty=\tau_0<\tau_1<...<\tau_{2^b}=\infty$, and the set of $2^b$ labels ${\cal L}_b=\{l_0,l_1,...,l_{2^b-1}\}$ where $l_p \in (\tau_p,\tau_{p+1}]$, for $p \in [0,2^b-1]$ \cite{jacobsson2017throughput}. Let us assume that ${\mathbf x}_k\sim \mathcal{CN}({\mathbf 0},\,{\mathbf R}_{x_k})$, where ${\mathbf R}_{x_k} \in \mathbb{C}^{M \times M}$ is the covariance matrix of ${\mathbf x}_k$. We now use Bussgang's theorem \cite{bussgang1952crosscorrelation} to derive a model for the quantized vector ${\mathbf x}_{k,Q}$, which we will use later to derive our DQA-LMS algorithm. Employing Bussgang's theorem, ${\mathbf x}_{k,Q}$ can be decomposed as
\begin{align}
    {\mathbf x}_{k,Q}={\mathbf G}_{k,b}{\mathbf x}_k+{\mathbf q}_k, \vspace{-0.25em}\vspace{-0.25em}\label{dcmps}
\end{align}
where the quantization distortion ${\mathbf q}_k$ is uncorrelated with ${\mathbf x}_k$, and ${\mathbf G}_{k,b} \in \mathbb{R}^{M \times M}$ is a diagonal matrix described by 
\begin{equation}
    \begin{split}
   \vspace{-0.25em}
   {\mathbf G}_{k,b} = {\rm diag}({\mathbf R}_{x_k})^{-\frac{1}{2}} \sum_{j=0}^{2^b-1} & \frac{l_j}{\sqrt{\pi}} \left[\exp(-\tau_j^2 {\rm diag} ({\mathbf R}_{x_k})^{-1}) \right.\\
        & \left.-\exp(-\tau_{j+1}^2 {\rm diag} ({\mathbf R}_{x_k})^{-1})\right].
    \end{split}  \label{Gk}\raisetag{25pt}
\end{equation}
Note that this signal decomposition is also applied to the desired signal, $d_{k,Q}$, which is the output of the second ADC in the system, and for the particular case that ${\mathbf R}_{x_k}= \mathbb{E}[{\mathbf x}_k{\mathbf x}_k^H] = \sigma_{x,k}^2 {\mathbf I}_M$, ${\mathbf G}_{k,b}$ becomes $g_{k,b} {\mathbf I}_M$. However, to minimize the mean square error (MSE) between ${\mathbf x}_k$ and ${\mathbf x}_{k,Q}$, we need to characterize the probability density function (PDF) of ${\mathbf x}_k$ to find the optimal quantization labels. Since the choice of labels based on the PDF is not practical, we assume the regressor ${\mathbf x}_{k}(i)$ is Gaussian,
adapt the approach in \cite{jacobsson2017throughput} and approximate the thresholds and labels as follows:
\begin{enumerate}
  \item We generate an auxiliary Gaussian random variable with unit variance and then use the Lloyd-Max algorithm \cite{lloyd1982least}, \cite{max1960quantizing} to find a set of thresholds ${\cal {\widetilde{T}}}_b=\{\tau_1,\dots,\tau_{2^b-1}\}$ and labels ${\cal {\widetilde{L}}}_b=\{\widetilde{l}_0,\dots,\widetilde{l}_{2^b-1}\}$ that minimize the MSE between the unquantized and the quantized signals.
  
  \item We complete the set of thresholds ${\cal T}_b$ by adding $\tau_0=-\infty$ and $\tau_{2^b}=\infty$ to the set ${\cal {\widetilde{T}}}_b$.
  
  \item We rescale the labels such that the variance of the auxiliary random variable is 1. To do this, we multiply each label in the set ${\cal {\widetilde{L}}}_b$ by \vspace{-0.75em}
  \begin{equation}
  \alpha = \Big(2 \sum \limits_{j=0}^{2^b-1} \tilde{l}_j^2(\Phi (\sqrt{2\tau_{j+1}^2})-\Phi (\sqrt{2\tau_{j}^2})\Big)^{-1/2}
      %\alpha = \frac{1}{\sqrt{2 \sum\limits_{j=0}^{2^b-1} \tilde{l}_j^2(\Phi (\sqrt{2\tau_{j+1}^2})-\Phi (\sqrt{2\tau_{j}^2}))}}
\vspace{-0.75em}
\end{equation}
to produce a set of suboptimal labels ${\cal L}_b = \alpha {\cal {\widetilde{L}}}_b$ , where $\Phi(.)$ is the cumulative distribution function (CDF) of a standard Gaussian random variable.
\end{enumerate}

We generate these thresholds and labels offline to build  ${\mathbf G}_{k,b}$ for the proposed DQA-LMS algorithm in what follows.
%\vspace{-0.8em}

\subsection{Derivation of DQA-LMS}\label{AA}
\label{ssec:derv}
We consider ${\mathbf x}_k(t)$ and $d_k(t)$ as the analog input and output of the unknown system ${\mathbf w}_o$ at node $k$. Let ${\mathbf x}_k(i)$ and $d_k(i)$ denote the high-precision sampled versions of ${\mathbf x}_k(t)$ and $d_k(t)$, and ${\mathbf x}_{k,Q}(i)$ and $d_{k,Q}(i)$ denote the coarsely quantized versions of ${\mathbf x}_k(i)$ and $d_k(i)$, respectively. We assume that the input signal at each node is  Gaussian with zero mean and covariance matrix ${\mathbf R}_{x_k}=E[{\mathbf x}_k {\mathbf x}_k^H]= \sigma_{x,k}^2 {\mathbf I}_M$ for $k=1,2,...,N$. Using \eqref{dcmps}, we can decompose ${\mathbf x}_{k,Q}(i)$ and $d_{k,Q}(i)$ as
\begin{gather}
   {\mathbf x}_{k,Q}(i) = g_{k,b}(i){\mathbf x}_k(i)+{\mathbf q}_{x,k}(i),\\
   \begin{split}   
   d_{k,Q}(i) &= Q(d_k(i)) \approx g_{k,b}(i)d_k(i)+q_k(i) \\
    %\qquad &= g_{k,b}(i)( {\mathbf w}_o^H{\mathbf x}_k(i)+v_{k}(i))+q_k(i) \\
    \qquad &= g_{k,b}(i){\mathbf w}_o^H{\mathbf x}_k(i)+\hat{q}_k(i) \label{dcmpsD},
\end{split}
\end{gather}
where $\hat{q}_k(i)=g_{k,b}(i)v_{k}(i)+q_k(i)$ and $g_{k,b}(i)$ are built from an estimate of ${\mathbf R}_{x_k}$ given by $\widehat{{\mathbf R}}_{x_k} ={\mathbf x}_k{\mathbf x}_k^H$ \cite{li2017channel} that depends on the choice of  ${\mathbf x}_k$ due to \eqref{regres}. Because the adaptive algorithm receives a quantized signal, ${\mathbf x}_{k,Q}$, and the signal is assumed to be wide-sense stationary, at each time instant, we estimate $\sigma_{x,k}^2$ using the variance of the received input, $\sigma_{x_{k,Q}}^2$ and the distortion factor of the $b$-bit quantization, $\rho_{k,b}$, such that $\sigma_{x,k}^2 \approx \sigma_{x_{k,Q}}^2 + \rho_{k,b}$, where $\rho_{k,b} \approx \frac{\pi \sqrt{3}}{2} 2^{-2b}$  \cite{mezghani2007modified} for a Gaussian signal using non-uniform quantization to obtain the scalar $g_{k,b}(i)$.

We show next that a learning algorithm based directly on \eqref{dcmpsD} is biased for estimating $\mathbf{w}_o$, and show how to correct for this bias.  For this, let $\beta_k(i)$ be a coefficient to be chosen shortly, and define $\hat{d}_k(i) = \beta_k(i){\mathbf w}_{k}^H(i-1){\mathbf x}_{k,Q}(i)$ and construct an MSE cost function as described by
\begin{equation}
    \begin{split}
        J_k({\mathbf w}_{k}(i))  &= \mathbb{E}[| e_{k,Q}(i)|^2] =  \mathbb{E}[|d_{k,Q}(i) - \hat{d}_k(i)|^2] \\
        & =  \mathbb{E}[|d_{k,Q}(i) - \beta_k(i){\mathbf w}_{k}^H(i-1){\mathbf x}_{k,Q}(i)|^2]  , \label{jk2}
    \end{split}
\end{equation}
which depends only on the observed quantized quantities $d_{k,Q}(i)$ and $\mathbf{x}_{k,Q}(i)$. 
For $\beta_k(i) = 1$ as in DLMS, the quantization of $d_k(i)$ would result in biased estimates of $\mathbf{w}_o$. 
In the following we show how to optimally choose $\beta_k(i)$ to reduce the bias. The proposed gradient-descent recursion to perform distributed learning based on \eqref{jk2} is described by
\begin{align}
    {\mathbf h}_{k}(i)={\mathbf w}_{k}(i-1)-\mu_{k}\nabla J_k({\mathbf w}_{k}(i-1)). \label{hk}
\end{align}
To compute the gradient of \eqref{hk}, we write the error in \eqref{jk2} as
 \begin{equation}
     \begin{split}
        e_{k,Q}(i) &= d_{k,Q}(i)-\beta_k(i){\mathbf w}_{k}^H(i-1){\mathbf x}_{k,Q}(i)\\
       % &= g_{k,b}(i){\mathbf w}_o^H{\mathbf x}_k(i)+\hat{q}_k(i)-\beta_k(i){\mathbf w}_{k}^H(i-1){\mathbf x}_{k,Q}(i)\\
        &= g_{k,b}(i){\mathbf w}_o^H{\mathbf x}_k(i)+\hat{q}_k(i)-\beta_k(i){\mathbf w}_{k}^H(i-1)\\
        &\qquad (g_{k,b}(i){\mathbf x}_k(i)+{\mathbf q}_{x,k}(i))\\
  &=g_{k,b}(i)({\mathbf w}_o^H - \beta_k(i){\mathbf w}_{k}^H(i-1)) {\mathbf x}_k(i) \\
        &\qquad -\beta_k(i){\mathbf w}_{k}^H(i-1){\mathbf q}_{x,k}(i)+\hat{q}_k(i). \label{erq}
     \end{split}
 \end{equation}
We assume that ${\mathbf R}_{x_k} = \mathbb{E}[{\mathbf x}_k{\mathbf x}_k^H] = \sigma_{x,k}^2 {\mathbf I}_M$ and ${\mathbf R}_{q,k} = \mathbb{E}[{\mathbf q}_{x,k}{\mathbf q}_{x,k}^H] = \sigma_{q,k}^2 {\mathbf I}_M$. Substituting \eqref{erq} in \eqref{jk2} and taking the expected value of \eqref{hk}, we have
\begin{equation}
    \begin{split}
        \mathbb{E} [{\mathbf h}_{k}(i)] = &[{\mathbf I}_M-\mu_{k} g_{k,b}^2(i)\beta_k(i) {\mathbf R}_{x_k} -\mu_{k} g_{k,b}(i)\beta_k(i) {\mathbf R}_{q,k}]\\
        &\cdot\mathbb{E} [{\mathbf w}_{k}(i-1)]+\mu  g_{k,b}^2(i) {\mathbf R}_{x_k} {\mathbf w}_o^H. \label{exph}
    \end{split}\raisetag{10pt}
\end{equation}
Substituting the values of $\mathbf{R}_{x_k}$ and $\mathbf{R}_{q,k}$ and taking the limit on \eqref{exph}, we obtain
\begin{equation}
    %\begin{split}
        \lim_{i \to +\infty}  \mathbb{E} [{\mathbf h}_{k}(i)] %&= (g_{k,b}(i) {\mathbf R}_{x,k}+ {\mathbf R}_{q,k})^{-1} 
        %\frac{g_{k,b}(i)}{\beta_k(i)} {\mathbf R}_{x_k} {\mathbf w}_o\\ &
        = \frac{1}{\beta_k(i)}\frac{g_{k,b}(i)\sigma_{x,k}^2}{g_{k,b}(i) \sigma_{x,k}^2+\sigma_{q,k}^2}{\mathbf w}_o.
    %\end{split}\raisetag{17pt}
\end{equation}
We conclude that the solution is unbiased if we choose
\begin{equation}
    \beta_k(i) = \frac{g_{k,b}(i)\sigma_{x,k}^2}{g_{k,b}(i) \sigma_{x,k}^2+\sigma_{q,k}^2}. \label{eq:beta} 
\end{equation}
 The gradient of $|e_{k,Q}(i)|^2$ with respect to ${\mathbf w}_k^H$ is $\nabla J_k({\mathbf w}_{k}(i-1))=-\frac{g_{k,b}(i)\sigma_{x,k}^2}{g_{k,b}(i) \sigma_{x,k}^2+\sigma_{q,k}^2}{\mathbf x}_{k,Q}(i) e_{k,Q}^*(i)$. After organizing the terms of the gradient, we obtain the DQA-LMS algorithm: 
\begin{equation}
\begin{split}
    {\mathbf h}_{k}(i) &= {\mathbf w}_{k}(i-1)+\mu_{k}\frac{g_{k,b}(i)\sigma_{x,k}^2}{g_{k,b}(i) \sigma_{x,k}^2+\sigma_{q,k}^2}{\mathbf x}_{k,Q}(i)e_{k,Q}^*(i), \\
    {\mathbf w}_{k}(i) &= \sum _{l\in \mathcal {N}_{k}}a_{lk}{\mathbf h}_{l}(i),  \vspace{-0.5em} \label{qwkd}
%\vspace{-0.75em}
\end{split}\raisetag{20pt}
\end{equation}
%where
%$\sigma_{x,k}^2 \approx \sigma_{x_{k,Q}}^2 + \rho_{k,b}$, $e_{k,Q}(i)=d_{k,Q}(i)-\frac{g_{k,b}(i)\sigma_{x,k}^2}{g_{k,b}(i) \sigma_{x,k}^2 + \sigma_{q,k}^2}{\mathbf w}_{k}^H(i-1){\mathbf x}_{k,Q}(i)$ and $g_{k,b}(i) = \frac{1}{\sqrt{\sigma_{x_k}^2}} \sum\limits_{j=0}^{2^b-1} \frac{l_j}{\sqrt{\pi}} [\exp(\frac{-\tau_{j}^2}{\sigma_{x_k}^2}) -\exp(\frac{-\tau_{j+1}^2}{\sigma_{x_k}^2})] $.
\begin{equation}
    \begin{split}
       \vspace{-0.5em} &e_{k,Q}(i)=d_{k,Q}(i)-\frac{g_{k,b}(i)\sigma_{x,k}^2}{g_{k,b}(i) \sigma_{x,k}^2+\sigma_{q,k}^2}{\mathbf w}_{k}^H(i-1){\mathbf x}_{k,Q}(i) \,,\\
        &g_{k,b}(i) = \frac{1}{\sqrt{\sigma_{x_k}^2}} \sum\limits_{j=0}^{2^b-1} \frac{l_j}{\sqrt{\pi}} [\exp(\frac{-\tau_{j}^2}{\sigma_{x_k}^2}) -\exp(\frac{-\tau_{j+1}^2}{\sigma_{x_k}^2})], \, %\\
        %&{\rm and}\ \, \sigma_{x,k}^2 \approx \sigma_{x_{k,Q}}^2 + \rho_{k,b}.
    \end{split}\raisetag{22pt}
\end{equation}
and $\sigma_{x,k}^2 \approx \sigma_{x_{k,Q}}^2 + \rho_{k,b}$. The scalar $g_{k,b}$ can be computed offline when ${\mathbf R}_{x,k}$ is known and wide-sense stationary and must be estimated online when ${\mathbf R}_{x,k}$ is unknown or non-stationary.
\vspace{-1em}
\subsection{Computational Complexity and Energy Consumption}\label{AB}
\label{ssec:compx}
Table \ref{CC} shows the computational complexity of the DQA-LMS algorithm in terms of the number of multiplications and additions at node $k$ per time instant, where $n_k$ is the number of neighbor nodes connected to node $k$. At each time instant, DQA-LMS performs a few more operations ($\approx O(2^b)$) than DLMS. Note that we compute $g_{k,b}(i)$ online since this is more appropriate for non-stationary input data. However, one can compute ${\mathbf G}_{k,b}$ offline if an estimate of ${\mathbf R}_{x_k}$ in \eqref{Gk} is available.

\begin{figure}[htbp]
    \centering
    \includegraphics[width=6.25cm]{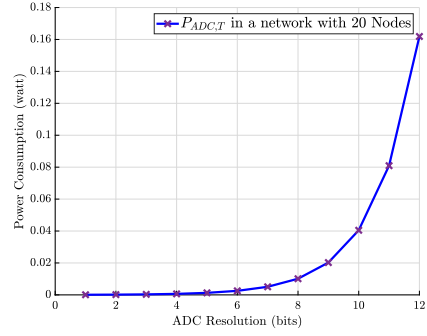}
    \vspace{-0.75em}
    \caption{Power consumption of the ADCs in an adaptive IoT network.}
    \vspace{-1.75em}
    \label{Padc1}
\end{figure}

However, the extra complexity of DQA-LMS allows the system to work in a more energy-efficient way. In order to assess the power savings by low resolution quantization, we consider a network with $N$ nodes in which each node uses two ADCs. The power consumption of each ADC is $P_{ADC}(b) = cB 2^b$ \cite{orhan2015low}, where $B$ is the bandwidth (related to the sampling rate), $b$ is the number of quantization bits of the ADC, and $c$ is the power consumption per conversion step. Therefore, the total power consumption of the ADCs in the network is
\begin{equation}
    P_{ADC,T}(b) = 2 N cB 2^b \qquad  ({\rm watts}). \label{Padcn}
\end{equation}
Fig.~\ref{Padc1} shows an example of the total power consumption of ADCs in a narrowband IoT (NB-IoT) network running diffusion adaptation consisting of 20 nodes with bandwidth $B=200 \, {\rm kHz}$ \cite{ratasuk2016nb} and considering the power consumption per conversion step of each ADC, $c=494~ {\rm fJ}$, as in \cite{chung20097}.

\vspace{-0.25em}
\begin{table*}[htb!]
\begin{footnotesize}
\caption{Computational Complexity per Time Instant}
\vspace{-1.5em}
\begin{center}
\begin{tabular}{|c|c|c|c|c|}
\hline
\textbf{Task} & \textbf{{Multiplications}}& \textbf{{Additions}} & \textbf{{Divisions}} & \textbf{{Exponentiations}} \\
\hline
$g_{k,b}(i) = \frac{1}{\sqrt{\sigma_{x_k}^2}} \sum\limits_{j=0}^{2^b-1} \frac{l_j}{\sqrt{\pi}} [\exp(\frac{-\tau_{j}^2}{\sigma_{x_k}^2}) -\exp(\frac{-\tau_{j+1}^2}{\sigma_{x_k}^2})]$ & $2^{b+1}+1$ & $2^b-1$ & $2^b+1$ & $2^b$ \\
\hline
$\beta_k(i) = \frac{g_{k,b}(i)\sigma_{x,k}^2}{g_{k,b}(i) \sigma_{x,k}^2+\sigma_{q,k}^2}$ & $2$ & $1$ & $1$ & $0$ \\
\hline
$\hat{d}_{k,Q}(i) = \beta_k(i){\mathbf w}_{k}^H(i){\mathbf x}_{k,Q}(i)$ & $M+1$ & $M-1$  & $0$ & $0$ \\
\hline
$e_{k,Q}(i)=d_{k,Q}(i)-\hat{d}_{k,Q}(i)$ & $0$ & $1$ & $0$ & $0$ \\
\hline
${\mathbf h}_{k}(i+1)={\mathbf w}_{k}(i)+\mu_{k}\beta_k(i)e_{k,Q}^*(i){\mathbf x}_{k,Q}(i)$ & $M+2$ & $M$ & $0$ & $0$ \\
\hline
${\mathbf w}_{k}(i+1)=\sum\limits_{l\in \mathcal {N}_{k}}a_{lk}{\mathbf h}_{l}(i+1)$& $n_k M$ & $n_k M$  & $0$ & $0$ \\
\hline
\textbf{Total} (at node $k$) & $(2+n_k)M+2^{b+1}+6 $ & $(2+n_k)M+2^b $ & $2^b+2$ & $2^b$ \\
\hline
\end{tabular}
\label{CC}
\end{center}
\end{footnotesize}
\vspace{-2.5em}
\end{table*}

\section{Analysis of DQA-LMS}
\label{sec:perf}
In this section, we find sufficient conditions for all local estimates to converge in the mean to the unknown parameter vector ${\mathbf w}_o$ by using the evolution of the weight error vectors \cite{lopes2008diffusion}. Let us consider the global quantities of the network: ${\mathbf W}_o \triangleq [{\mathbf w}_o,\dots, {\mathbf w}_o]_{(NM \times 1)}$, ${\mathbf d}_Q(i) \triangleq [d_{1,Q}(i),\dots, d_{N,Q}(i)]^T$, $        {\mathbf v}_i \triangleq [v_1(i),\dots, v_N(i)]^T$,         ${\mathbf X}_Q(i) \triangleq {\rm diag} [{\mathbf x}_{1,Q}^T(i),\dots, {\mathbf x}_{N,Q}^T(i)]$.
%\begin{equation}
%    \begin{split}
%        {\mathbf W}_o &\triangleq [{\mathbf w}_o,\dots, {\mathbf w}_o]_{(NM \times 1)}\\
%        {\mathbf d}_Q(i) &\triangleq [d_{1,Q}(i),\dots, d_{N,Q}(i)]^T, \\
%        {\mathbf v}_i &\triangleq [v_1(i),\dots, v_N(i)]^T, \\
%        {\mathbf X}_Q(i) &\triangleq {\rm diag} [{\mathbf x}_{1,Q}^T(i),\dots, {\mathbf x}_{N,Q}^T(i)].
%    \end{split}
%\end{equation}

Using these quantities, the global form of \eqref{regres} is given by
%\begin{equation}
    ${\mathbf d}_Q(i) = {\mathbf W}_o^H {\mathbf X}_Q(i)+{\mathbf v}(i)$.% \label{regresglob}. %\vspace{-0.25em}
%\end{equation}
Defining $\mathbf{B}(i)$, $\mathbf{W}(i)$ and $\mathbf{H}(i)$ as the global quantities for, respectively, $\beta_k(i)$, $\mathbf{h}_k(i)$ and $\mathbf{w}_k(i)$, we can express \eqref{qwkd} as
\begin{equation}
    \begin{split}
        {\mathbf H}(i) &= {\mathbf W}(i-1)+{\mathbf D}{\mathbf B}(i){\mathbf X}_Q(i)({\mathbf d}_Q(i)\\
        &-{\mathbf B}(i){\mathbf W}^H(i-1){\mathbf X}_Q(i))^*, %\\
       \quad  {\mathbf W}(i) = {\mathbf C} {\mathbf H}(i), %\vspace{-0.5em} \label{qwkdglb} 
    \end{split}\raisetag{10pt}
\end{equation}
which can be written in a compact form as
\begin{equation}\begin{split}
    {\mathbf W}(i) &= {\mathbf C} {\mathbf W}(i-1)+{\mathbf C} {\mathbf D}{\mathbf B}(i){\mathbf X}_Q(i)({\mathbf d}_Q(i)\\
    &-{\mathbf B}(i){\mathbf W}^H(i-1){\mathbf X}_Q(i))^*, \end{split}\label{wglobcomp}
\end{equation}
where ${\mathbf D} \triangleq {\rm diag} \{\mu_1 I_M,\dots,\mu_N I_M \}$ and ${\mathbf C}$ is an $MN \times MN$ matrix based on the combination coefficients, $a_{lk}$, defined as
\begin{align}
        {\mathbf A} &\triangleq \begin{bmatrix} a_{ij} %a_{11} & \dots & a_{1N} \\ \vdots & \ddots & \vdots \\ a_{N1} & \dots & a_{NN}
        \end{bmatrix}, & 
        {\mathbf C} &\triangleq {\mathbf A} \otimes {\mathbf I}_M. \vspace{-0.25em}
    %\end{split}
\end{align}

Using the independence assumption \cite{lopes2008diffusion} that states that ${\mathbf x}_{k,Q}(i)$ and $v_k(i)$ are
i.i.d. in time and space with $\sigma_{v,k}^2 = \mathbb{E}[|v_k(i)|^2]$, and  $v_k(i)$ is independent of ${\mathbf x}_{k,Q}(i)$, we define the weight error vector, $\widetilde{{\mathbf w}}_k(i)$ and its global vector $\widetilde{{\mathbf w}}(i)$ as
\begin{equation}
    \widetilde{{\mathbf W}}(i) \triangleq {\mathbf W}_o - {\mathbf W}(i). \vspace{-0.25em}
\end{equation}
Note that using diffusion combination policies for $a_{lk}$, we have ${\mathbf C}{\mathbf W}_o = {\mathbf W}_o$ \cite{lopes2008diffusion}. Subtracting ${\mathbf W}_o$ from the left-hand side and ${\mathbf C}{\mathbf W}_o$ from the right-hand side of \eqref{wglobcomp}, we have
\begin{equation}
    \begin{split}
        \vspace{-0.25em}
        \widetilde{{\mathbf W}}(i) &= {\mathbf C}{\mathbf W}_o - {\mathbf C} {\mathbf W}(i-1) - {\mathbf C} {\mathbf D}{\mathbf B}(i){\mathbf X}_Q(i)\\
        %& \qquad ({\mathbf d}_Q(i)-{\mathbf B}(i){\mathbf W}^H(i-1){\mathbf X}_Q(i))^*\\
        %&= {\mathbf C}{\mathbf W}_o - {\mathbf C} {\mathbf W}(i-1) - {\mathbf C} {\mathbf D}{\mathbf B}(i){\mathbf X}_Q(i)\\
        %& \qquad ({\mathbf W}_o^H {\mathbf X}_Q(i)+{\mathbf v}(i)-{\mathbf B}(i){\mathbf W}^H(i-1){\mathbf X}_Q(i))^*\\
        &= {\mathbf C}\widetilde{{\mathbf W}}(i-1) - {\mathbf C} {\mathbf D}{\mathbf B}(i){\mathbf X}_Q(i)\\
        & \qquad ({\mathbf X}_Q^*(i){\mathbf B}(i)^*\widetilde{{\mathbf W}}(i-1)+{\mathbf v}^*(i))\\
        &= {\mathbf C}({\mathbf I}_{MN}-{\mathbf D}{\mathbf B}(i){\mathbf X}_Q(i){\mathbf X}_Q^*(i){\mathbf B}(i)^*)\widetilde{{\mathbf W}}(i-1) \\
        & \qquad - {\mathbf C} {\mathbf D}{\mathbf B}(i){\mathbf X}_Q(i){\mathbf v}^*(i).%\vspace{-0.5em} 
        \label{statespace}\raisetag{10pt}
    \end{split}
\end{equation}
Taking the expectation of both sides of \eqref{statespace}, we have
\begin{equation}
    \mathbb{E} [\widetilde{{\mathbf W}}(i)] = C({\mathbf I}_{MN}-{\mathbf D}{\mathbf R}_Q)\mathbb{E} [\widetilde{{\mathbf W}}(i-1)], \label{exptilde}
\end{equation}
where $C \triangleq \mathbb{E} [{\mathbf C}]$, ${\mathbf R}_Q \triangleq {\rm diag} \{{\mathbf R}_{1,Q},\dots,{\mathbf R}_{N,Q}\}$ and ${\mathbf R}_{k,Q}= \mathbb{E} [{\mathbf B}_k(i){\mathbf x}_{k,Q}(i){\mathbf x}_{k,Q}^*(i){\mathbf B}_k(i)^*]$.

To ensure stability of the recursion in \eqref{exptilde} with the independence assumption and using combinations that satisfy \eqref{alk}, there exist sufficiently small step-sizes $\mu_k < \mu_{max}$ such that
\begin{equation}
        \|\mathbb{E} [\widetilde{{\mathbf W}}(i)]\|_{bl_\infty} \leq \|C\|_{bl_\infty}.\|{\mathbf E}_i\|_{bl_\infty}. \|\mathbb{E} [\widetilde{{\mathbf w}}(i-1)]\|_{bl_\infty}, \label{expinfty}
\end{equation}
where $\|.\|_{bl_\infty}$ denotes the block maximum norm \cite{takahashi2010diffusion} and ${\mathbf E}_i={\mathbf I}_{MN} - {\mathbf D}{\mathbf R}_Q$. In order for DQA-LMS to converge, we hold \eqref{expinfty} such that $\|{\mathbf E}\|_{bl_\infty}<1$ and $\|C_i\|_{bl_\infty}\leq 1$ for all $i \geq 0$. It is proven in \cite{takahashi2010diffusion} that possibly random, time-varying convex combinations generated by ATC or CTA diffusion algorithms ensure $\|C_i\|_{bl_\infty}\leq 1$. Therefore, to find sufficient conditions on step-sizes, we must have ${\mathbf I}_{MN} - {\mathbf D}{\mathbf R}_Q<1$.

We now employ the eigenvalue decomposition ${\mathbf R}_{k,Q}= {\mathbf \Phi}_{k,Q} {\mathbf \Lambda}_{k,Q} {\mathbf \Phi}_{k,Q}^H$, where ${\mathbf \Lambda}_{k,Q}$ is an $M \times M$ diagonal matrix consisting of the eigenvalues $\{\lambda_{(k,Q)_1},\dots,\lambda_{(k,Q)_M}\}$ of ${\mathbf R}_{k,Q}$, and the matrix ${\mathbf \Phi}_{k,Q}$ is an $M \times M$ square matrix whose columns are the eigenvectors $\{\boldsymbol{\phi}_{(k,Q)_1},\dots,\boldsymbol{\phi}_{(k,Q)_M}\}$ of ${\mathbf R}_{k,Q}$ associated with these eigenvalues. We define ${\mathbf \Phi}_Q \triangleq {\rm diag} \{ {\mathbf \Phi}_{1,Q}, \dots, {\mathbf \Phi}_{N,Q} \}$ and ${\mathbf \Lambda}_Q \triangleq {\rm diag} \{ {\mathbf \Lambda}_{1,Q}, \dots, {\mathbf \Lambda}_{N,Q} \}$. Since ${\mathbf \Phi}_Q^H {\mathbf D} {\mathbf \Phi}_Q = {\mathbf D}$, the condition on the step size can be written as $\|{\mathbf I}_{MN} - {\mathbf D}{\mathbf \Lambda}_Q\|_\infty < 1$, which yields 
\begin{equation}
    \begin{split}
        \|{\mathbf I}_{MN} - {\mathbf D} {\mathbf \Lambda}_Q\|_\infty & = \max_{1 \le k \le N} \|{\mathbf I}_M - \mu_k {\mathbf \Lambda}_{k,Q}\| \\
        & = \max_{1 \le k \le N} \max_{1 \le m \le M} |1 - \mu_k \lambda_{(k,Q)_m}| <1, \nonumber \vspace{-0.25em}
    \end{split}
\end{equation}
where $\lambda_{(k,Q)_m}$ is the $m$th diagonal eigenvalue of ${\mathbf R}_{k,Q}$. Therefore, the stability condition for DQA-LMS is given by
\begin{equation}
    0 < \mu_k < \frac{2}{\lambda_{\max}({\mathbf R}_{k,Q})} \quad {\rm for\ all}\ k = 1,2,\dots,N. \vspace{-0.75em}
\end{equation}

\section{Simulation Results}
\label{sec:sims}
\begin{figure}[htbp]
    \centering
    \hspace{-0.75em}\subfloat[Distributed network structure\label{1a}]
        {\includegraphics[width=4.75cm]{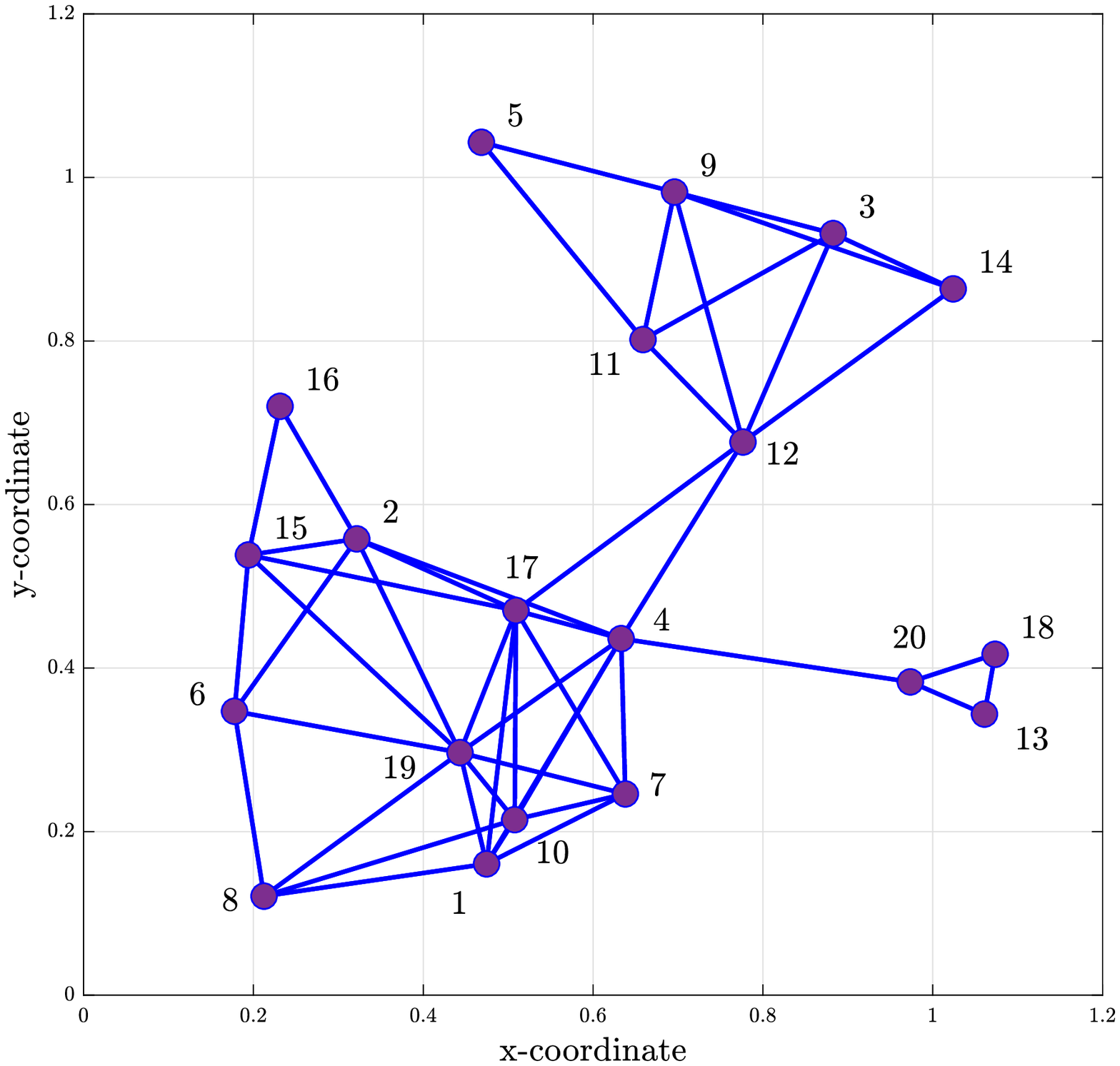}}
    \hfill
    \hspace{-1.25em}\subfloat[Variances and correlation coefficients\label{1b}]
        {\includegraphics[width=4.75cm]{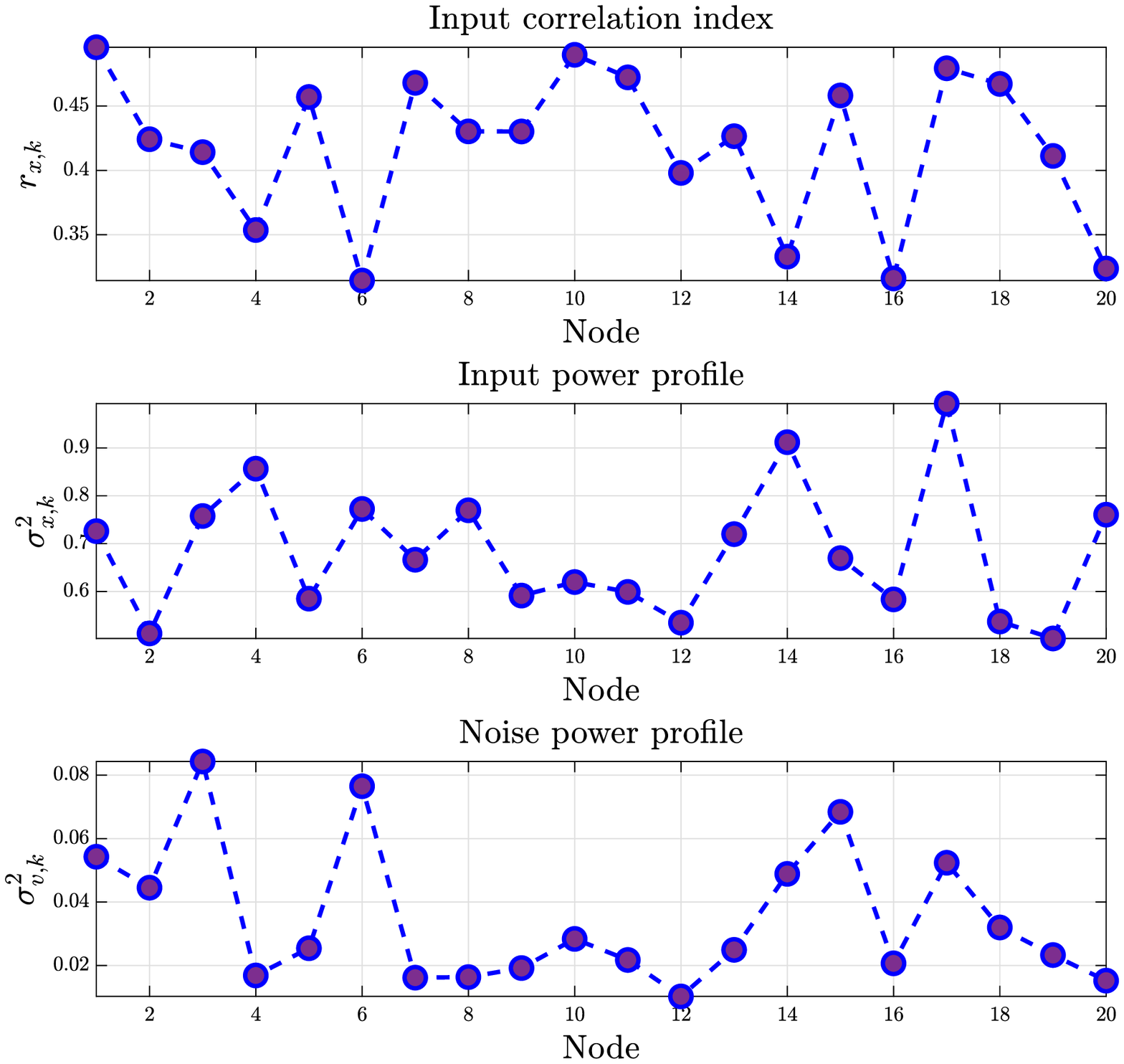}}
        \vspace{-0.5em}
    \caption{A wireless network with $N=20$ nodes.}
    \label{netw}
\end{figure}

In this section, we assess the performance of the DQA-LMS algorithm for a parameter estimation problem in an IoT network with $N=20$ nodes. The impulse response of the unknown system has $M=8$ taps, is generated randomly and normalized to one. The input signals ${\mathbf x}_k(i)$ at each node are generated by passing a white Gaussian noise process with variance $\sigma_{x,k}^2$ through a first order autoregressive model with transfer function $ \frac{1}{1-r_{x,k} z^{-1}}$ where $ r_{x,k} \in (0.3,0.5)$ are the correlation coefficients and quantized using Lloyd-Max quantization scheme to generate ${\mathbf x}_{k,Q}(i)$. The noise samples of each node are drawn from a zero mean white Gaussian process with variance $\sigma_{v,k}^2$. Fig. ~\ref{netw} plots the network details.

The simulated mean-square deviation (MSD) learning curves are obtained by ensemble averaging over 100 independent trials. We choose the same step sizes for all agents, i.e., $\mu_k=0.05$. The combining coefficients $a_{lk}$ are computed by the Metropolis rule. The evolution of the ensemble-average learning curves, $\frac{1}{N}\mathbb{E}[\|\widetilde{{\mathbf w}}_i\|^2]$, for the ATC diffusion strategy using different numbers of bits is assessed. The theoretical MSD of the DLMS with the same step size $\mu$ and the Metropolis rule applied to $a_{lk}$ is approximated by $\frac{\mu M}{N^2} \sum_{k=1}^N {\sigma_{v,k}^2}$ \cite{sayed2013diffusion} and shown by curve 1. Curve 2 shows the standard DLMS performance assuming full resolution ADCs to perform system identification. Curves 3, 5 and 7 show the MSD evolution of the standard DLMS with low resolution signals coarsely quantized with 1, 2 and 3 bits, respectively. Curves 4, 6 and 8 show the MSD performance of the proposed DQA-LMS algorithm that improves the error measurement confronted with coarsely quantized signals. The performance of the proposed DQA-LMS algorithm is closer to the DLMS while it reduces about $90\%$ of the power consumption by ADCs in the network (see Fig.~\ref{Padc1}). \vspace{-1.0em}
\begin{figure}[htbp]
\centerline{\includegraphics[width=7.85cm]{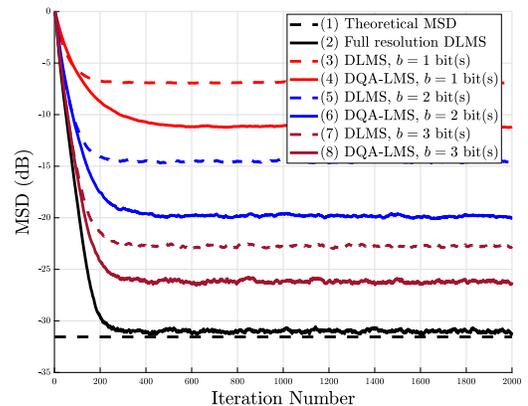}}
\vspace{-1.0em}
\caption{The MSD curves for the DLMS and DQA-LMS algorithms.}
\vspace{-1.0em}
\label{msd}
\end{figure}
\vspace{-0.25em}
\section{Conclusion}
\label{sec:conc}

In this paper, we have proposed an energy-efficient framework for distributed learning and developed the DQA-LMS algorithm using low-resolution ADCs for adaptive IoT networks. DQA-LMS has comparable computational cost to the full-resolution DLMS algorithm while it enormously reduces the power consumption of the ADCs in the network. Simulations have shown the close performance of DQA-LMS to the DLMS algorithm despite dealing with coarsely quantized signals.

% References should be produced using the bibtex program from suitable
% BiBTeX files (here: strings, refs, manuals). The IEEEbib.bst bibliography
% style file from IEEE produces unsorted bibliography list.

\bibliographystyle{IEEEbib}
\bibliography{refs}

\end{document}